\documentclass{article}

\usepackage{microtype}
\usepackage{graphicx}
\usepackage{subcaption}
\usepackage{booktabs} % for professional tables
\usepackage{booktabs} % for professional tables
\usepackage{balance}
\usepackage{enumitem}

\newcommand{\ie}{\emph{i.e.}}

\usepackage{hyperref}

\usepackage[accepted]{icml2026}

\usepackage{amsmath}
\usepackage{amssymb}
\usepackage{mathtools}
\usepackage{amsthm}
\usepackage{bm}

\usepackage{times}
\usepackage{epsfig}
\usepackage{multirow}
\usepackage{bbold}
\usepackage{threeparttable}
\usepackage{array}

\usepackage{bbding}
\usepackage{makecell}
\usepackage{balance}  
\usepackage{diagbox}
\usepackage{colortbl}
\usepackage{pifont}

\newlength\secmargin
\newlength\subsecmargin
\newlength\paramargin
\newlength\figmargin
\newlength\eqmargin
\definecolor{Gray}{rgb}{0.5, 0.5, 0.5}  
\definecolor{tableblue}{RGB}{204, 204, 255}
\definecolor{rowgray}{gray}{0.9} % 定义浅灰色背景
\definecolor{figpurple}{RGB}{146, 112, 202} 
\definecolor{figorange}{RGB}{225, 145, 60}
\definecolor{figgreen}{RGB}{118, 170, 115}
\definecolor{trajorange}{RGB}{230,126,34}
\definecolor{trajblue}{RGB}{41,128,185}

\usepackage[capitalize,noabbrev]{cleveref}

\theoremstyle{plain}

\theoremstyle{definition}

\theoremstyle{remark}

\usepackage{xcolor}

\usepackage[textsize=tiny]{todonotes}

\icmltitlerunning{Mitigating Error Accumulation in Continuous Navigation
via Memory-Augmented Kalman Filtering}

\begin{document}

\twocolumn[
  % \icmltitle{Rectifying State Drift in Aerial Vision-and-Language Navigation \\ via Non-Parametric Memory Correction}
  \icmltitle{Mitigating Error Accumulation in Continuous Navigation \\ via Memory-Augmented Kalman Filtering}




\icmlsetsymbol{projectlead}{\ensuremath{*}}
\icmlsetsymbol{corresponding}{\ensuremath{\dagger}}

\begin{icmlauthorlist}
    \icmlauthor{Yin Tang}{csu1,cityu}
    \icmlauthor{Jiawei Ma}{cityu,projectlead}
    \icmlauthor{Jinrui Zhang}{csu}
    \icmlauthor{Alex Jinpeng Wang}{csu}
    \icmlauthor{Deyu Zhang}{csu,corresponding}
\end{icmlauthorlist}

  \icmlaffiliation{csu1}{Big Data Institute, Central South University, Changsha, China. Work done while working at CityUHK as a visiting scholar.}
  \icmlaffiliation{cityu}{Department of Computer Science \& Institute of Digital Medicine, City University of Hong Kong, Hong Kong, China}
  \icmlaffiliation{csu}{School of Computer Science, Central South University, Changsha, China}
  \icmlcorrespondingauthor{Deyu Zhang}{zdy876@csu.edu.cn}


  \vskip 0.3in
]



% \printAffiliationsAndNotice{}
\printAffiliationsAndNotice{\textsuperscript{*}Project Leader.}

\begin{abstract}
Continuous navigation in complex environments is critical for Unmanned Aerial Vehicle (UAV).
However, the existing Vision-Language Navigation (VLN) models follow the dead-reckoning, which iteratively updates its position for the next waypoint prediction, and subsequently construct the complete trajectory. Then, such stepwise manner will inevitably lead to accumulated errors of position over time, resulting in misalignment between internal belief and objective coordinates, which is known as ``state drift'' and ultimately compromises the full trajectory prediction.
Drawing inspiration from classical control theory, we propose to correct for errors by formulating such sequential prediction as a \textbf{recursive Bayesian state estimation} problem. In this paper, we design \textbf{NeuroKalman}, a novel framework that decouples navigation into two complementary processes: a \textit{Prior Prediction}, based on motion dynamics and a \textit{Likelihood Correction}, from historical observation. We first mathematically associate Kernel Density Estimation of the measurement likelihood with the attention-based retrieval mechanism, which then allows the system to rectify the latent representation using retrieved historical anchors without gradient updates. 
Comprehensive experiments on TravelUAV benchmark demonstrate that, with only 10\% of the training data fine-tuning, our method clearly outperforms strong baselines and regulates drift accumulation. The code will be publicly released at \url{https://github.com/yinntag/Neuro-Kalman}.

\end{abstract}

%\vspace{-2mm}
\section{Introduction}
%\vspace{-2mm}
\label{sec:intro}
% \alex{Highlight contribution/evidences/etc. for all the paper}
Continuous navigation is fundamental to achieving full automation in Unmanned Aerial Vehicles (UAVs)~\cite{liu2023aerialvln, lee2025citynav}. Recent advance has been made to utilize Vision-Language Navigation (VLN) models, which follow a global natural language instruction and the local visual observation at the current step, to predict the next waypoint~\cite{fan2023aerial, wang2024towards, gao2025openfly}. By incrementally updating the position from past estimates for the next forecasting, the model predicts the full trajectory for UAV navigation step-by-step, which draws a parallel to the concept of dead-reckoning in control theory.
% sequentially forecasting waypoints step-by-step, the model predicts the full trajectory that  draws a parallel to the concept of dead-reckoning in control theory, \ie, the UAV navigates by . 

\begin{figure}[t!]
	\centering
	\includegraphics[width=1\linewidth]{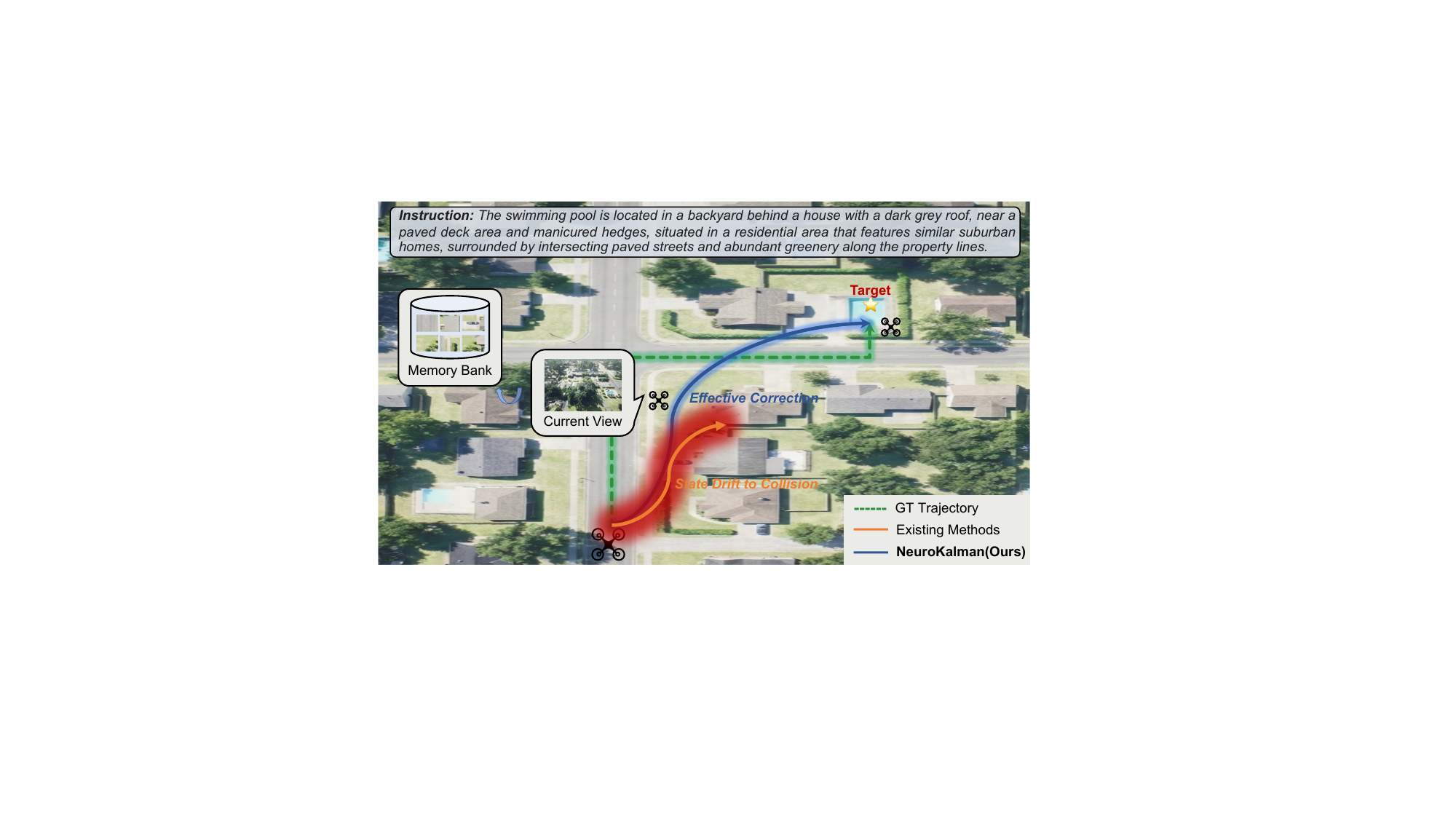}
	%\vspace{-4mm}
    \caption{\textbf{Illustration of state drift mitigation.} Given a global instruction, existing models ignore the history but make prediction only from current inputs, and thus suffer from accumulated error and state drift to collision (\textit{\textcolor{trajorange}{orange line}}). Instead, our NeuroKalman framework introduces a Kalman correction mechanism by fusing historical measurements as anchors for prediction to rectify the trajectory prediction (\textit{\textcolor{trajblue}{blue line}}).
    % \alex{the instruction font is too small}
    }
    \label{fig:motivation}
	%\vspace{-4mm}
\end{figure}

Nevertheless, such mechanisms are sensitive to individual outlier predictions~\cite{wang2024towards, wang2024vision, cai2025flightgpt} and will accumulative error over time. As illustrated in Figure \ref{fig:motivation}, without a dedicated correction mechanism, the disparity between the UAV position and the expected trajectory intensifies. As the linguistic instruction remains unchanged throughout the navigation~\cite{liu2023aerialvln}, it is then assumed to already indicate a global plan implicitly from the initial position to the target destination. Then, the disparity, between the objective UAV coordinates and the expected (internal) positional belief, a concrete instantiation to the concept of ``state drift'', will cause a fundamental misalignment in the latent space between current visual observation and the initial language instruction for subsequent waypoints prediction, which further magnify the error in navigation and hurts model generalization~\cite{krantz2020beyond, georgakis2022cross, chen2022think}.

% in position when the UAV deviates, will also caused  will also lead to a misalignment between the  

 % In detail, the accumulation of per-step errors may lead to significant position (state) deviation. As shown in Figure~\ref{fig:motivation}, given a linguistic instruction at the beginning, the model may already have an internal model over states on how to plan the trajectory. However, as the error accumulates, without a dedicated correction mechanism, the disparity between the internal positional belief and its objective global coordinates intensifies. Then, the internal model, guided by the global language instructions that remain unchanged, cannot correctly realize the change of state in reality. Since the language instructions remain unchanged, this divergence leads to a fundamental misalignment between the command and the actual state, which further magnify the error in subsequent predictions~\cite{krantz2020beyond, georgakis2022cross, chen2022think}.

%  As illustrated in Figure \ref{fig:motivation}, the accumulation of per-step errors leads to significant drift,' where the model's internal positional belief diverges from objective global coordinates. Because the global linguistic instructions remain static, the model fails to reconcile its internal state with the actual physical position. This fundamental misalignment between the unchanging command and the drifting state further compounds errors in subsequent trajectory predictions.

Draw inspiration from classical Bayesian state estimation~\cite{sarkka2023bayesian}, we consider a recursive feedback loop where the historical measurements are used to periodically correct internal predictions. In detail, it consists of two complementary phases: a \textit{Prediction} step that generates a prior estimate based on motion dynamics, and an \textit{Update} step that rectifies this belief using the likelihood of historical measurements. We note that this diverges from existing models~\cite{wang2024towards, lin2025openvln} which primarily focus on generating strong prior estimation while overlooking an explicit mechanism to calibrate these predictions~\cite{kloss2021train, revach2022kalmannet}.

% exhibit a structural imbalance: while they possess strong predictive capabilities (\ie, a prior), they lack the explicit mechanism to calibrate these predictions against observation evidence (\ie, the likelihood).

% \alex{add some linking words/connectors. For example, driven by our inspiration}
To this end, we propose NeuroKalman, which directly models the Bayesian prediction-update cycle among the latent representations, and decomposes the navigation into two streams. For the \textit{Prediction}, we employ a Recurrent Neural Network (RNN) to model the motion dynamics for initial forecasting. 
Then, with the mathematical association between Kernel Density Estimator (KDE) over the measurement likelihood~\cite{katharopoulos2020transformers} and the attention mechanism~\cite{vaswani2017attention}, we introduce an Memory Bank to integrate historical observation in the \textit{Update}, which can be jointly used to estimate Kalman Gain to refine the prediction and consequently mitigate the accumulated errors in navigation.

% We demonstrate that attention-based retrieval over memory banks is mathematically equivalent to a  Finally, by fusing the GRU's prediction with the memory-augmented observation via a learned Kalman Gain, the navigation system is allowed to obtain a rectified output (\emph{i.e.,} a reliable posterior) that is immune to state drift.

Compared with conventional navigation models~\cite{chen2021history, chen2022think}, \textbf{our prediction–update NeuroKalman architecture offers a unique advantage in preventing overfitting}. Existing models typically requires massive datasets to learn a generalized transition function, which are still prone to overfitting when data diversity is limited. In contrast, NeuroKalman retrieves relevant historical visual observations. This allows it to effectively utilize the necessary information to regularize VLN model training by enforcing temporal smoothness. In this way, we align the semantics of multiple local observations with global instructions, improving the next waypoint prediction.
% . This allows our method to ``anchor'' the agent in the true state manifold even with minimal fine-tuning. 
Experiments on TravelUAV benchmark~\cite{wang2024towards} show that NeuroKalman, fine-tuned on only 10\% of the training data, significantly outperforms the initial model and the naive finetuning baselines. Our contributions can be summarized as follows:
\begin{itemize}[leftmargin=*]
    \item We point out the accumulation of error in continuous UAV navigation, and investigates a correction mechanism based on historical content and temporal smoothness for robust VLN model prediction \& generalization. 
    \item We formulate the navigation as a recursive Bayesian state estimation problem and propose the corresponding NeuroKalman framework. With the mathematical association between KDE of the likelihood function and the attention-based memory retrieval, we model the Bayesian prediction-update cycle and integrate the selected memory of historical observation to correct the prediction and mitigate state drift in continuous environments.    
    % a novel UAV-VLN architecture that explicitly models the Bayesian prediction-update cycle. By fusing a parametric GRU with observations calibrated by non-parametric memory via a learnable Kalman Gain, our model effectively 
    \item With a randomly sampled subset of training data for model finetuning, our method clearly improve the model performance on the TravelUAV benchmark with a clear margin. This also verifies its effectiveness in mitigating overfitting when finetuning over limited data.
    % We demonstrate that NeuroKalman achieves state-of-the-art performance on the TravelUAV benchmark using 10\% training data finetuing, significantly surpassing baselines trained on the full dataset. 
\end{itemize}

%\vspace{-2mm}
\section{Related Works}
%\vspace{-1mm}
\label{sec:rw}

\noindent\textbf{Vision-and-Language Navigation (VLN).} 
VLN research has evolved significantly from indoor environments to large-scale aerial scenarios~\cite{anderson2018vision, jain2019stay}. Early approaches like AerialVLN~\cite{liu2023aerialvln} and CMA~\cite{anderson2018vision} operated on discrete graphs, selecting actions from pre-defined steps~\cite{krantz2020beyond}. Some methods rely on large-scale pre-training or data augmentation to improve generalization~\cite{guhur2021airbert, chen2021history}. However, this setting is overly simplified and fails to match realistic UAV dynamics. Recent works, such as CityNav~\cite{lee2025citynav} and TravelUAV~\cite{wang2024towards}, have shifted to continuous environments to better simulate real-world flights. OpenVLN~\cite{lin2025openvln} introduces a data-efficient approach for continuous control, while NavFoM~\cite{zhang2025embodied} proposes a unified foundation model capable of processing multimodal inputs across varying horizons. Despite these advancements, existing methods typically rely on static networks with fixed weights. They lack the ability to correct accumulated errors during long flights, leading to the state drift problem.

\noindent\textbf{Temporal Context Modeling.} 
Temporal context modeling is crucial for long-horizon navigation, where the agent needs to incorporate historical context for trajectory planning to avoid getting lost. Early methods used recurrent networks like LSTMs~\cite{anderson2018vision} or GRUs~\cite{chung2014empirical} to encode history, but they suffer from information loss over long horizons~\cite{fried2018speaker}. Later works introduced explicit memory structures to extend the context window. For example, MapNet~\cite{henriques2018mapnet} and Transformer-XL~\cite{dai2019transformer} store historical states, while recent methods like SkyVLN~\cite{li2025skyvln}, OpenFly~\cite{gao2025openfly}, and CityNavAgent~\cite{zhang2025citynavagent} utilize topological maps or key-frames to assist reasoning. However, these approaches typically treat memory as a passive buffer, simply aggregating historical features with current observations. In contrast, we adopt a ``retrieve-to-correct'' paradigm~\cite{khandelwal2019generalization, borgeaud2022improving}. Instead of simple feature concatenation, we use retrieved memory as probabilistic evidence to explicitly rectify the agent's belief state through Bayesian fusion, actively correcting potential drift.

\noindent\textbf{Deep Bayesian Filtering \& State Estimation.} 
Addressing the distribution shift in deployment environments has recently popularized Test-Time Adaptation (TTA) techniques in embodied AI~\cite{wang2020tent, kumar2021rma}. Recent methods like FEEDTTA~\cite{kim2025test} and FSTTA~\cite{gao2023fast} attempt to mitigate drift via feedback-based reinforcement learning or online gradient updates. However, in UAV-VLN, the lack of reliable supervision often causes these methods to reinforce existing errors~\cite{niu2022efficient}. Alternatively, Deep Bayesian Filtering combines neural networks with probabilistic models~\cite{kloss2021train, revach2022kalmannet}. While effective at learning transition prior dynamics, these approaches struggle to define a robust likelihood for high-dimensional visual inputs, often reverting to parametric models that are themselves prone to drift~\cite{haarnoja2016backprop, becker2019recurrent}. In contrast, NeuroKalman addresses this by framing navigation as Recursive Bayesian Estimation. Instead of updating model weights like TTA, we correct the belief state. By deriving the likelihood directly from memory retrieval, our framework offers a stable solution to state drift.

%\vspace{-2mm}
\section{Method}
%\vspace{-1mm}
\label{sec:our}
\begin{figure*}[t!]
	\centering
	\includegraphics[width=\linewidth]{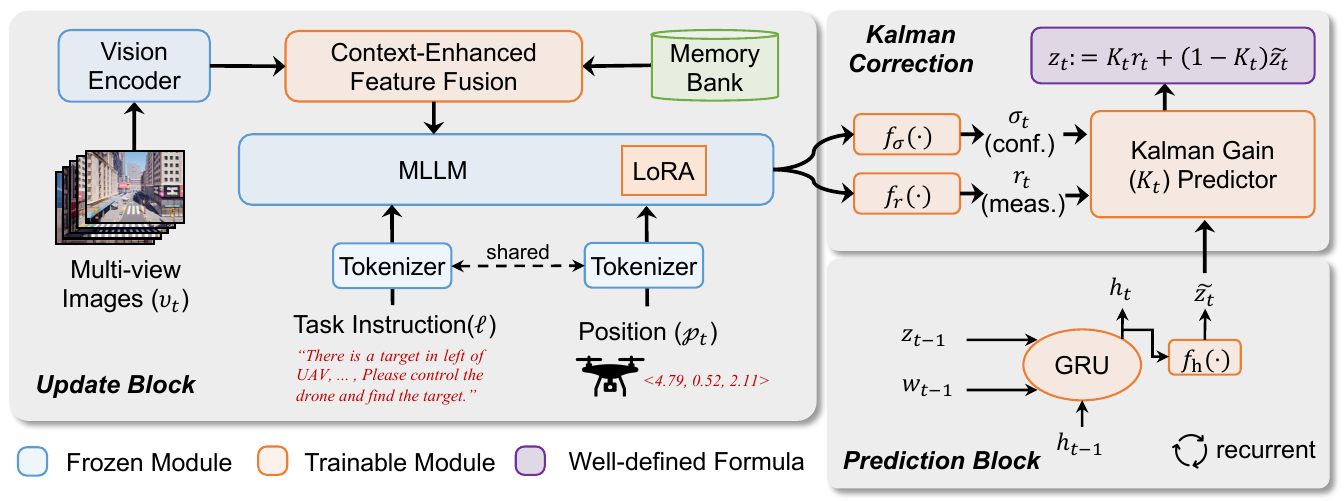}
	%\vspace{-2mm}
	\caption{\textbf{NeuroKalman framework} aims to leverage temporal context to enhance next step prediction in navigation. Specifically, we follow the logic in classic Kalman filtering~\cite{sarkka2023bayesian}, and consider the \textit{Prediction} and \textit{Update} steps~\cite{kalman1960new}, \ie, the former one makes initial estimation while the latter one estimates measurement representation $\mathbf{r}_t$ for core Kalman correction. In detail, the \textit{Prediction Block} employs a GRU to roughly model the motion dynamics to predict the prior state $\tilde{\mathbf{z}}_t$ with updated hidden state $\mathbf{h}_t$, according to the posterior state $\mathbf{z}_{t-1}$ in the last step. Then, with the confidence scalar $\sigma_t$ predicted by the \textit{Update Block}, the Kalman Gain $K_t$ is estimated on the representation space for correction. The waypoint prediction $\phi(\mathbf{z}_t)$ is omitted for the clarity of illustration while the variables $\mathbf{r}_t$, $\tilde{\mathbf{z}}_t$ can be both fed in $\phi(\cdot)$ for augmented supervision.}
    % the architecture decouples navigation into . Specifically, the \textit{Prediction Block} employs a GRU to model the agent's dynamics, utilizing the  and waypoint displacement $\mathbf{w}_t$ to estimate a  while updating the temporal hidden context . Simultaneously, the 
    % \textit{Update Block} leverages an MLLM backbone to encode multimodal inputs into a measurement feature $\mathbf{r}_t$ along with an uncertainty confidence score $\sigma$. These streams are fused via the \textit{Kalman Correction} module to produce a robust posterior state $\mathbf{z}_t$ for precise waypoint prediction. 
	\label{fig:pipeline}
	%\vspace{-4mm}
\end{figure*}

To address the error accumulation inherent in dead-reckoning \cite{hong2021vln, chen2021history}, we depart from the conventional paradigm that treats navigation purely as a sequential prediction task~\cite{anderson2018vision, fan2023aerial}, but instead reframe it as a recursive Bayesian state estimation problem. As illustrated in Figure \ref{fig:pipeline}, this formulation enables us to structurally decouple the \textit{Prediction Block}, for initial prior estimation, from the \textit{Update Block} for refinement from historical measurements, transforming navigation from a blind rollout into a reliable inference process. Accordingly, we posit that robust navigation relies on maintaining a probabilistic belief over the state space. This is achieved by fusing the motion dynamics prior with reliable measurements via the \textit{Kalman Correction} to obtain a rectified posterior that prevents drift.

\subsection{Problem Formulation: Navigation as Filtering}
Formally, we consider a UAV operating in a continuous 3D environment \cite{krantz2020beyond, wang2024towards}. At each time step $t$, the model receives an observation tuple $o_t = \{v_t, p_t, l\}$, comprising multi-view visual inputs $v_t$ (\ie, front, back, down, left, right), current 3D coordinates $p_t$, and the global natural language instruction $l$. The model predicts a next waypoint $w_t$ for UAV's execution. Central to our formulation is the high-dimensional latent belief state $\mathbf{z}_t \in \mathbb{R}^d$. Unlike simple waypoint coordinates, $\mathbf{z}_t$ encodes a semantic understanding of the UAV's position and environmental context from time step $t$. Our core objective is to estimate the posterior distribution of this state, $P(\mathbf{z}_t | o_{1:t}, w_{1:t-1})$, given the entire history of observations and waypoints. Under the standard Markov assumption, this posterior estimation decomposes into a recursive prediction-update cycle, known as the Bayes filter \cite{sarkka2023bayesian}:
\begin{equation}
    \underbrace{P(\mathbf{z}_t | o_{1:t}, w_{1:t-1})}_{\text{Posterior}} \propto \underbrace{P(o_t | \mathbf{z}_t)}_{\text{Likelihood}} \times \underbrace{P(\mathbf{z}_t | \mathbf{z}_{t-1}, w_{t-1})}_{\text{Prior}}
    \label{eq:bayes_filter}
\end{equation}

Since these distributions are intractable in high-dimensional visual spaces, we propose NeuroKalman to structurally instantiate this logic within a neural architecture \cite{kloss2021train, haarnoja2016backprop}. As detailed in the following sections, we decouple Eq.~\ref{eq:bayes_filter} into three learnable modules:
\begin{itemize}[leftmargin=*]
    \item Predictive Prior (Section~\ref{sec:prediction}): A RNN-based predictor that models the transition $P(\mathbf{z}_t | \mathbf{z}_{t-1}, w_{t-1})$, serving as the dead-reckoning mechanism.
    \item Measurement Likelihood (Section~\ref{sec:update}): An MLLM that encodes multi-modal inputs with historical memory to parameterize the likelihood $P(o_t | \mathbf{z}_t)$, providing retrieved evidence as corrective anchors.
    \item The Kalman Correction (Section~\ref{sec:correction}): A gating mechanism that dynamically fuses the Prior and Likelihood to compute the rectified Posterior $P(\mathbf{z}_t | o_{1:t}, w_{1:t-1})$.
\end{itemize}

%%%%%%%%%%%%%%%%%%%%%%%%%%%%%%%%%%%%%%%%%%%%%%%%%%%%%%%%%%%%%%%%%%%%%%%%%%%%%%%%%%%%
\subsection{The Prediction Step: Predictive Prior}
\label{sec:prediction}
The Prediction step is to primarily estimate the \textit{a prior} belief of the current state, functioning as an internal motion model, based solely on learned transition dynamics. This step effectively serving as a dead-reckoning mechanism for the Bayesian filter, and we design an RNN with Gated Recurrent Unit (GRU) to parameterize the transition distribution $P(\mathbf{z}_t | \mathbf{z}_{t-1}, w_{t-1})$. 

Formally, at time step $t$, this module processes three inputs: the previous posterior $\mathbf{z}_{t-1}$, which serves as the rectified belief state implicitly encoding the UAV's optimal position at $t-1$; the previous waypoint $\mathbf{w}_{t-1}$, defined as the UAV's displacement vector in the body coordinate system from $t-1$ to $t$; and the hidden state $\mathbf{h}_{t-1}$, which encodes the historical motion dynamics accumulated from the initial step up to $t-1$. To initialize the recursive process at $t=0$, we set the initial state $\mathbf{z}_0$ to the measurement output of the MLLM at the initial position (denoted as $\mathbf{r}_0$). The initial hidden state $\mathbf{h}_0$ is then obtained by mapping $\mathbf{z}_0$ through a MLP layer with a Tanh activation to align with the GRU's latent space. Based on these inputs, the GRU updates its internal hidden state and projects the prior estimate $\tilde{\mathbf{z}}_t$ as follows:
\begin{align}
    \mathbf{h}_t &= \text{GRU}([\mathbf{z}_{t-1}, \mathbf{w}_{t-1}], \mathbf{h}_{t-1}) \\
    \tilde{\mathbf{z}}_t &= \text{MLP}_{prior}(\mathbf{h}_t)
\end{align}
Here, $\tilde{\mathbf{z}}_t$ denotes the prior prediction st time step $t$. Crucially, this prediction is a ``blind'' process: it is derived purely from dead-reckoning logic without accessing the current visual observation $v_t$~\cite{banino2018vector}. While the GRU effectively captures temporal dependencies and smooths the trajectory, it remains a purely parametric model governed by fixed weights. In unseen scenarios during training or data-scarce regimes, reliance on this motion dynamic prediction $\tilde{\mathbf{z}}_t$ inevitably leads to error accumulation, causing the latent belief to drift from the true manifold. This necessitates the subsequent likelihood correction step~\cite{liu2023aerialvln, wang2024towards}.
%%%%%%%%%%%%%%%%%%%%%%%%%%%%%%%%%%%%%%%%%%%%%%%%%%%%%%%%%%%%%%%%%%%%%%%%%%%%%%%%%%%%%%
\subsection{The Update Step: Measurement Likelihood}
\label{sec:update}
As the Prediction step primarily uses motion dynamics for prior estimation, we implement the \textit{Update Block} to fuse the \textit{historical memory} functioning as external information, with the current observations and correspondingly implement a Multimodal Large Language Model (MLLM) to parameterize the measurement likelihood. At each time step $t$, we first augment the current visual input $v_t$ with relevant historical context retrieved from the episodic memory bank $\mathcal{M}$. The MLLM then jointly processes the memory-augmented visual features, the global instruction $l$, and the position of the UAV $p_t$ to output the latent measurement $\mathbf{r}_t$ along with a confidence score $\sigma_t \in [0, 1]$ representing the measurement uncertainty. This $\mathbf{r}_t$ serves as the measurement likelihood for the subsequent Kalman correction.

\textbf{Memory Construction.} To facilitate the likelihood estimation, the memory bank $\mathcal{M}$ is constructed incrementally to store high-fidelity historical visual anchors. We adopt a post-correction storage strategy~\cite{yang2024samurai, shi2025memoryvla}: the visual representation corresponding to the waypoint localization decoded from the \textit{rectified posterior state} is appended to $\mathcal{M}$ only if the system's confidence score exceeds a reliability threshold (\emph{i.e.,} $\sigma_t > 0.5$).
Formally, the bank is defined as $\mathcal{M} = \{(\mathbf{k}_i, \mathbf{v}_i)\}_{i=1}^N$, where $\mathbf{k}_i = \mathbf{v}_i$ encodes the fixed visual features of selected snapshots. This selective storage mechanism ensures that the MLLM always attends to ``validated anchors''~\cite{chen2022think}, preventing the accumulation of noisy or ambiguous measurements.
%%%%%%%%%%%%%%%%%%%%%%%%%%%%%%%%%%%%%%%%%%%%%%%%%%%%%%%%%%%%%%%%%%%%%%%%%%%%%%%%%%%%%

\textbf{Retrieval as Kernel Density Estimation.} 
We rigorously formulate the memory retrieval process not merely as feature matching, but as Kernel Density Estimation (KDE) over the visual feature manifold. Given that our retrieval operates purely within the visual domain, where the query is the current visual feature $\mathbf{f}_t$ and the memory stores historical visual features $\{\mathbf{f}_i\}_{i=1}^N$, our goal is to derive a refined visual evidence $\hat{\mathbf{z}}_{evi}$. From a statistical perspective, we treat the historical features as empirical samples drawn from the underlying observation distribution. We employ the Nadaraya-Watson kernel regression estimator to approximate the expected canonical feature $\mathbb{E}[\mathbf{f} | o_t]$ on the visual manifold~\cite{nadaraya1964estimating, watson1964smooth}:
\begin{equation}
    \hat{\mathbf{z}}_{evi} = \frac{\sum_{i=1}^N \mathcal{K}(\mathbf{f}_t, \mathbf{f}_i) \cdot \mathbf{f}_i}{\sum_{j=1}^N \mathcal{K}(\mathbf{f}_t, \mathbf{f}_j)}
    \label{eq:nadaraya_watson}
\end{equation}
where $\mathbf{f}_t$ serves as the query, $\mathbf{f}_i$ serves as both key and value, and $\mathcal{K}(\cdot, \cdot)$ is a kernel function measuring visual similarity. In our NeuroKalman architecture, we utilize the scaled dot-product attention mechanism. Defining the kernel as $\mathcal{K}(\mathbf{x}, \mathbf{y}) = \exp(\frac{\mathbf{x}^\top \mathbf{y}}{\sqrt{d}})$, Eq. \ref{eq:nadaraya_watson} becomes mathematically equivalent to the Softmax Attention operation~\cite{katharopoulos2020transformers, choromanski2020rethinking}:
\begin{equation}
    \alpha_i = \text{Softmax}\left(\frac{\mathbf{f}_t^\top \mathbf{f}_i}{\sqrt{d}}\right), \quad \hat{\mathbf{z}}_{evi} = \sum_{i=1}^N \alpha_i \mathbf{f}_i
\end{equation}
Through this derivation, the retrieved vector $\hat{\mathbf{z}}_{evi}$ functions as a measurement correction. The term $\alpha_i$ represents the posterior probability $P(\mathbf{f}_i | \mathbf{f}_t)$ that the current visual observation belongs to the same local manifold as history $i$. 
Consequently, $\hat{\mathbf{z}}_{evi}$ aggregates historical anchors to mitigate high-frequency visual noise before state estimation. Unlike the GRU prior which propagates uncertainty blindly, $\hat{\mathbf{z}}_{evi}$ leverages observational evidence to stabilize the likelihood term, ensuring robustness in out-of-distribution scenarios.

\subsection{The Kalman Correction: Fusion as Gain}
\label{sec:correction}
The final step fuses the potentially drifting prior estimate $\tilde{\mathbf{z}}_t$ (from Section \ref{sec:prediction}) with the encoded measurement $\mathbf{r}_t$  (from Section \ref{sec:update}) to obtain the final corrected posterior $\mathbf{z}_t$. We design a fusion mechanism that structurally mirrors the classic Kalman Filter update equation~\cite{kalman1960new}:
\begin{equation}
    \mathbf{z}_{post} = \mathbf{z}_{prior} + \mathbf{K}_t (\mathbf{y}_t - \mathbf{H}\mathbf{z}_{prior})
    \label{eq:classic_kalman}
\end{equation}
In our NeuroKalman framework, the measurement $\mathbf{y}_t$ corresponds to the encoded measurement $\mathbf{r}_t$. $\mathbf{z}_{prior}$ and $\mathbf{z}_{post}$ correspond to the prior estimate $\tilde{\mathbf{z}}_t$ and final corrected posterior $\mathbf{z}_t$, respectively. We assume an identity measurement matrix $\mathbf{H} = \mathbf{I}$ since both the prior and measurement are mapped into the same aligned latent feature space~\cite{haarnoja2016backprop}.

\textbf{Learnable Kalman Gain.}
The core of this correction is the optimal weighting between prior and measurement. Considering the intractability of explicitly estimating noise covariances (\emph{i.e.,} $\mathbf{Q}$ and $\mathbf{R}$) of traditional Kalman Filtering in high-dimensional latent space, we formulate a learnable gating network to approximate the Kalman Gain $\mathbf{K}_t$~\cite{revach2022kalmannet}. It dynamically computes the element-wise uncertainty based on the current context:
\begin{equation}
    \mathbf{K}_t = \text{Sigmoid}\left(\mathbf{W}_g [(\mathbf{r}_t - \tilde{\mathbf{z}}_t) \mathbin{;} \phi(\sigma_t)] + \mathbf{b}_g\right)
\end{equation}
where $[\cdot \mathbin{;} \cdot]$ denotes concatenation, $\phi(\cdot)$ is a learnable MLP projection mapping the confidence score $\sigma_t$ to the feature dimension, and $\text{Sigmoid}(\cdot)$ is the activation function ensuring the gain $\mathbf{K}_t \in (0, 1)^d$.

\textbf{Bayesian Update.}
Using the computed gain, the system updates the belief state. The fusion equation is defined as:
\begin{align}
    \mathbf{z}_t &= (1 - \mathbf{K}_t) \odot \tilde{\mathbf{z}}_t + \mathbf{K}_t \odot \mathbf{r}_t \label{eq:fusion} \\
    &= \tilde{\mathbf{z}}_t + \mathbf{K}_t \odot (\mathbf{r}_t - \tilde{\mathbf{z}}_t) \label{eq:residual_form}
\end{align}
Equation \ref{eq:residual_form} is algebraically identical to the standard Kalman correction form (Eq. \ref{eq:classic_kalman}), where $(\mathbf{r}_t - \tilde{\mathbf{z}}_t)$ represents the \textit{residual}—the difference between the external measurement and the predicted prior.

\begin{figure}[t!]
	\centering
	\includegraphics[width=\linewidth]{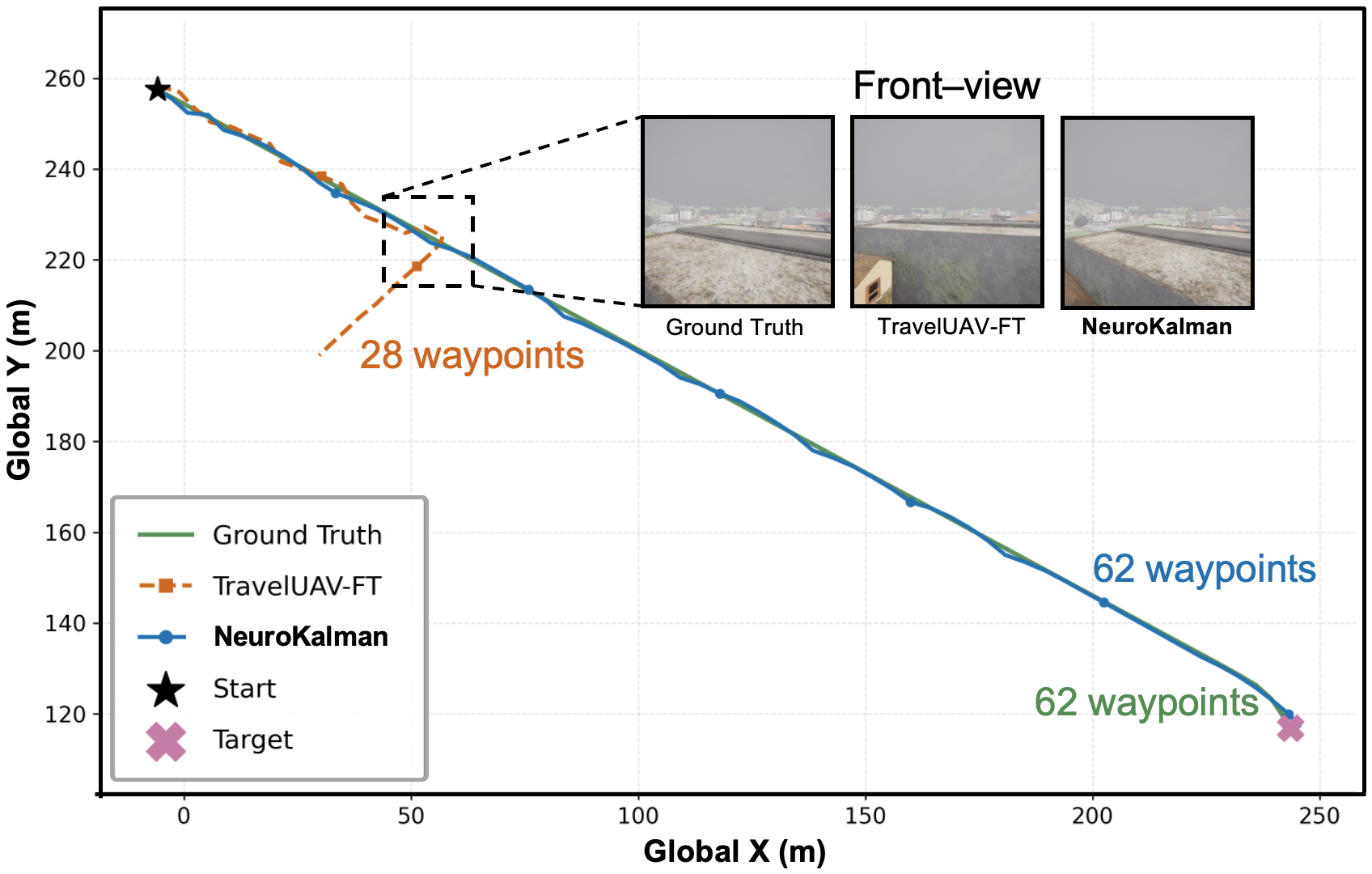}
	%\vspace{-4mm}
	\caption{
		\textbf{Demonstration of trajectory rectification.} The TravelUAV-FT relies solely on parametric predictions to estimate its trajectory, resulting in obvious trajectory drift. NeuroKalman rectifies its position by integrating Kalman correction.
	}
	\label{fig:drift}
	%\vspace{-4mm}
\end{figure}
\textbf{Dynamic Uncertainty Regulation.}
The learnable Kalman Gain $\mathbf{K}_t$ functions as an adaptive regulator. When the measurement is confident, the system heavily weighs the innovation to pull the state towards the validated measurement. Conversely, when the measurement is ambiguous, it relies more on the smooth internal dynamics of the Prior. This mechanism ensures that the final posterior $\mathbf{z}_t$ remains anchored to the true trajectory manifold, effectively cancelling the state drift. Finally, $\mathbf{z}_t$ is passed to the waypoint predictor to predict the next waypoint and fed back into the next prediction step as $\mathbf{z}_{t-1}$. As visualized in Figure \ref{fig:drift}, this correction mechanism actively performs micro-adjustments to re-align the belief state with the ground truth, effectively preventing the trajectory drift characteristic of the uncorrected baseline (TravelUAV-FT).

\begin{table*}[t]
	\centering
	\caption{Experimental results on the UAV-Need-Help test seen set, grouped by assistant levels (L1-L3). \textbf{Bold} indicates the best performance among all methods within each assistant level.}
	\label{tab:seen}
	
	% 1. 缩小列间距，防止文字被压缩太扁
	\setlength{\tabcolsep}{2pt}
	\renewcommand{\arraystretch}{.85}
	
	% 2. 使用 resizebox 强制适应宽度
	\resizebox{\textwidth}{!}{%
		\begin{threeparttable}
			\begin{tabular}{c|l|*{4}{c}|*{4}{c}|*{4}{c}}
				\toprule
				\multirow{2}{*}{\textbf{Assistant}}
				& \multirow{2}{*}{\textbf{Methods}}
				& \multicolumn{4}{c}{\textbf{Full}}
				& \multicolumn{4}{|c}{\textbf{Easy}} 
				& \multicolumn{4}{|c}{\textbf{Hard}} \\
				&
				& NE $\downarrow$& SR $\uparrow$& OSR $\uparrow$& SPL $\uparrow$
				& NE $\downarrow$& SR $\uparrow$& OSR $\uparrow$& SPL $\uparrow$
				& NE $\downarrow$& SR $\uparrow$& OSR $\uparrow$& SPL $\uparrow$ \\
				\midrule
				
				% ================= LEVEL 1 BLOCK =================
				\multirow{7}{*}{L1}
				& Random Action
				& 222.20 & 0.14 & 0.21 & 0.07        
				& 142.07 & 0.26 & 0.39 & 0.13        
				& 320.12 & 0.00 & 0.00 & 0.00\\
				& Fixed Action
				& 188.61 & 2.27 & 8.16 & 1.40			
				& 121.36 & 3.48 & 11.48 & 2.14 
				& 270.69 & 0.79 & 4.09 & 0.49\\
				& CMA~\cite{anderson2018vision}
				& 135.73 & 8.37 & 18.72 & 7.90
				& 84.89 & 11.48 & 24.52 & 10.68 
				& 197.77 & 4.57 & 11.65 & 4.51\\
				& TravelUAV~\cite{wang2024towards}    
				& 106.28 & 16.10 & 44.26 & 14.30
				& 68.78 & 18.84 & 47.61 & 16.39
				& 152.04 & 12.76 & 40.16 & 11.76\\
				& TravelUAV-FT~\cite{wang2024towards}    
				& 99.79 & 17.56 & 41.89 & 14.71
				& 64.10 & 20.69 & 45.98 & 16.79
				& 143.85 & 13.70 & 36.85 & 12.15\\
				& OpenVLN~\cite{lin2025openvln} 
				& 125.97 & 14.39 & 28.03 & 12.94 
				& 87.96 & 15.22 & 30.64 & 13.31
				& 175.54 & 13.32 & 24.62 & 12.55\\
				
				\rowcolor{rowgray} \cellcolor{white} 
				& \textbf{NeuroKalman (Ours)}
				& \textbf{71.56} & \textbf{25.86} & \textbf{58.73} & \textbf{22.43}
				& \textbf{42.70} & \textbf{30.52} & \textbf{62.70} & \textbf{25.86}
				& \textbf{105.07} & \textbf{20.11} & \textbf{53.90} & \textbf{18.21}\\
				\midrule

				% ================= LEVEL 2 BLOCK =================
				\multirow{5}{*}{L2}
				& CMA~\cite{anderson2018vision}
				& 141.55 & 7.02 & 15.39 & 6.54
				& 87.77 & 9.55 & 19.87 & 8.74 
				& 207.18 & 3.94 & 9.92 & 3.94\\
				& TravelUAV~\cite{wang2024towards} 
				& 120.57 & 12.98 & 37.38 & 11.30
				& 76.89 & 17.55 & 43.48 & 15.01 
				& 186.22 & 7.40 & 29.92 & 6.76\\
				& TravelUAV-FT~\cite{wang2024towards} 
				& 112.20 & 14.05 & 35.50 & 11.70
				& 72.50 & 18.80 & 42.00 & 15.50 
				& 178.00 & 8.20 & 27.50 & 7.50\\
				& OpenVLN~\cite{lin2025openvln} 
				& 129.68 & 13.83 & 25.97 & 12.18
				& 82.80 & 17.30 & 31.41 & 14.70
				& 178.32 & 10.24 & 20.31 & 9.57\\
				
				\rowcolor{rowgray} \cellcolor{white} 
				& \textbf{NeuroKalman (Ours)}
				& \textbf{87.11} & \textbf{22.32} & \textbf{53.68} & \textbf{19.40}			
				& \textbf{46.50} & \textbf{27.60} & \textbf{59.50} & \textbf{23.00} 
				& \textbf{138.00} & \textbf{15.80} & \textbf{46.50} & \textbf{14.50}\\
				\midrule

				% ================= LEVEL 3 BLOCK =================
				\multirow{5}{*}{L3}
				& CMA~\cite{anderson2018vision}
				& 140.93 & 4.89 & 11.56 & 4.41
				& 83.58 & 7.35 & 17.81 & 6.53 
				& 210.91 & 1.89 & 3.94 & 1.83\\
				& TravelUAV~\cite{wang2024towards} 
				& 146.32 & 6.31 & 15.39 & 5.10
				& 93.15 & 9.55 & 21.94 & 7.32
				& 215.85 & 2.36 & 7.40 & 2.17\\
				& TravelUAV-FT~\cite{wang2024towards} 
				& 135.00 & 4.51 & 12.06 & 3.70
				& 89.00 & 6.00 & 17.11 & 4.70
				& 198.78 & 2.68 & 5.83 & 2.47\\
				& OpenVLN~\cite{lin2025openvln} 
				& 146.42 & 2.70 & 6.96 & 2.19
				& 92.90 & 4.86 & 11.38 & 3.89
				& 201.96 & 0.47 & 2.36 & 0.43\\
				
				\rowcolor{rowgray} \cellcolor{white} 
				& \textbf{NeuroKalman (Ours)}
				& \textbf{107.55} & \textbf{8.89} & \textbf{17.28} & \textbf{7.01}			
				& \textbf{83.00} & \textbf{11.24} & \textbf{23.63} & \textbf{10.65} 
				& \textbf{179.07} & \textbf{5.98} & \textbf{9.45} & \textbf{4.89}\\
				\bottomrule
			\end{tabular}
		\end{threeparttable}%
	}
\end{table*}

%\vspace{-3mm}
\section{Experiments}
%\vspace{-2mm}
\label{sec:expr}
%\vspace{\subsecmargin}
%\subsection{Experimental Setup}
%\vspace{\subsecmargin}
In this section, we empirically validate the effectiveness of NeuroKalman on the challenging TravelUAV benchmark, with a specific focus on data efficiency and generalization. We first describe the experimental setup and evaluation protocol, then present main results, ablation studies, and a detailed analysis of state drift problem.
%%%%%%%%%%%%%%%%%%%%%%%%%%%%%%%%%%%%%%%%%%%%%%%%%%%%%%%%%%%%%%%%%%%%%%%%%%%%%%
\subsection{Experimental Setup}
We evaluate our framework on the TravelUAV benchmark~\cite{wang2024towards}, focusing on robust performance under data constraints. We detail the dataset, baselines, metrics, and our specific implementation protocols below.

\textbf{Dataset \& Simulation Environment.} We adopt the TravelUAV benchmark and its UAV-Need-Help dataset~\cite{wang2024towards}, simulated in the high-fidelity AirSim environment~\cite{shah2017airsim} which provides realistic physics and diverse outdoor conditions (urban, snowy, meadow, etc). The dataset contains 12,149 human-operated trajectories annotated with 89 object categories. Following the official protocol, the data is split as follows: the Training set contains 9,152 trajectories across 20 scenes; the Test-Seen set has 1,410 trajectories from training scenes; the Test-Unseen-Map set comprises 958 trajectories from 2 entirely novel scenes; and the Test-Unseen-Object set contains 629 trajectories targeting novel objects. Trajectories are categorized by length: ``Easy'' ($<250$m) and ``Hard'' ($\ge 250$m), with target distances ranging from 50 to 400 meters.

\textbf{Baselines.} We compare NeuroKalman against a comprehensive set of baselines:
(1) Random/Fixed Action: Naive models that select random poses or map instructions to fixed movements;
(2) CMA~\cite{anderson2018vision}: A standard bi-directional LSTM model using cross-modal attention;
(3) TravelUAV baseline~\cite{wang2024towards}: The official strong baseline provided by the benchmark;
(4) OpenVLN~\cite{lin2025openvln} and NavFoM~\cite{zhang2025embodied}: State-of-the-art approaches for continuous UAV navigation and multimodal foundation models.
Note that for the TravelUAV baseline, we report results under two settings: trained on full data (denoted as TravelUAV) and fine-tuned on 10\% training data (denoted as TravelUAV-FT).

\textbf{Metrics.} We report four standard metrics~\cite{anderson2018vision}: Navigation Error (NE), the average distance between the UAV’s final position and the target; Success Rate (SR), the percentage of episodes where the agent stops within 20 meters of the target; Oracle Success Rate (OSR), the success rate by accounting for the minimum distance to the target achieved at any point during the navigation trajectory; and Success weighted by Path Length (SPL), which balances success with trajectory efficiency.

\textbf{Implementation Details.}
We implement our model using PyTorch on 4$\times$NVIDIA RTX A6000 GPUs. We freeze the MLLM backbone (EVA-CLIP~\cite{sun2023eva} as the visual encoder and Vicuna-7B~\cite{chiang2023vicuna} as the language backbone) and only compute gradients for the visual projector, waypoint predictor, and LoRA~\cite{hu2022lora} layers using the Adam optimizer (Learning Rate=$5e-5$, Batch Size=16). We also use an additional L1 loss to simultaneously supervise both the predictive prior ($\tilde{\mathbf{z}}_t$) and measurement ($\mathbf{r}_t$), with a coefficient factor of 0.2. We adopt limited data fine-tuning for robustness evaluation. Specifically, both NeuroKalman and the TravelUAV are initialized with weights pre-trained on the full data, yet are subsequently fine-tuned using only a fixed random 10\% subset of the training data.

\begin{table*}[t]
	\centering
	\caption{Experimental results with the L1 assistant on the UAV-Need-Help test unseen set. \textbf{Bold} indicates the best results.}
	\label{tab:unseen}
	
	\setlength{\tabcolsep}{2pt} 
	\renewcommand{\arraystretch}{.85}
	
	\resizebox{\textwidth}{!}{%
		\begin{threeparttable}
			\begin{tabular}{c|l|*{4}{c}|*{4}{c}|*{4}{c}}
				\toprule
				\multirow{2}{*}{\textbf{Split}}
				& \multirow{2}{*}{\textbf{Methods}}
				& \multicolumn{4}{c}{\textbf{Full}}
				& \multicolumn{4}{|c}{\textbf{Easy}} 
				& \multicolumn{4}{|c}{\textbf{Hard}} \\
				& 
				& NE $\downarrow$& SR $\uparrow$& OSR $\uparrow$& SPL $\uparrow$
				& NE $\downarrow$& SR $\uparrow$& OSR $\uparrow$& SPL $\uparrow$
				& NE $\downarrow$& SR $\uparrow$& OSR $\uparrow$& SPL $\uparrow$ \\
				\midrule
				
				% ================= UO BLOCK =================
				\multirow{7}{*}{UO}
				& Random Action
				& 260.14 & 0.16 & 0.16 & 0.16        
				& 174.10 & 0.48 & 0.48 & 0.48        
				& 302.96 & 0.00 & 0.00 & 0.00\\
				& Fixed Action 
				& 212.84 & 3.66 & 9.54 & 2.16			
				& 151.66 & 6.70 & 13.88 & 3.72 
				& 243.29 & 2.14 & 7.38 & 1.38\\
				& CMA~\cite{anderson2018vision}
				& 155.79 & 9.06 & 16.06 & 8.68			
				& 102.92 & 14.83 & 22.49 & 13.90 
				& 182.09 & 6.19 & 12.86 & 6.08\\
				& TravelUAV~\cite{wang2024towards}    
				& 118.11 & 22.42 & 46.90 & 20.51			
				& 86.12 & 24.40 & 49.28 & 22.03 
				& 134.03 & 21.43 & 45.71 & 19.75\\
				& TravelUAV-FT~\cite{wang2024towards}    
				& 112.01 & 23.53 & 42.13 & 20.29
				& 64.80 & 34.45 & 54.07 & 28.97
				& 135.51 & 18.10 & 36.19 & 15.97\\
				& NavFoM~\cite{zhang2025embodied} 
				& 108.04 & 29.83 & 47.99 & 27.20			
				& 70.51 & 32.54 & 50.72 & 29.54 
				& 133.01 & \textbf{28.03} & 46.18 & \textbf{25.64}\\
				
				\rowcolor{rowgray} \cellcolor{white}
				& \textbf{NeuroKalman (Ours)}
				& \textbf{71.01} & \textbf{32.48} & \textbf{60.82} & \textbf{28.50}
				& \textbf{44.50} & \textbf{42.50} & \textbf{66.50} & \textbf{37.37}
				& \textbf{84.50} & 27.50 & \textbf{58.00} & 24.50\\
				\midrule
				
				% ================= UM BLOCK =================
				\multirow{7}{*}{UM}
				& Random Action
				& 202.98 & 0.00 & 0.00 & 0.00			
				& 158.46 & 0.00 & 0.00 & 0.00 
				& 265.88 & 0.00 & 0.00 & 0.00\\
				& Fixed Action 
				& 180.47 & 0.52 & 2.61 & 0.39			
				& 132.89 & 0.89 & 4.28 & 0.67 
				& 247.72 & 0.00 & 0.25 & 0.00\\
				& CMA~\cite{anderson2018vision}
				& 141.68 & 2.30 & 10.02 & 2.16			
				& 102.29 & 3.57 & 14.26 & 3.33 
				& 197.35 & 0.50 & 4.03 & 0.50\\
				& TravelUAV~\cite{wang2024towards}    
				& 138.80 & 4.18 & 20.77 & 3.84			
				& 102.94 & 4.63 & 22.82 & 4.24 
				& 189.46 & 3.53 & 17.88 & 3.28\\
				& TravelUAV-FT~\cite{wang2024towards}    
				& 117.84 & 4.68 & 19.03 & 3.17
				& 87.50 & 5.13 & 21.39 & 5.06
				& 160.79 & 4.05 & 15.69 & 3.32\\
				& NavFoM~\cite{zhang2025embodied} 
				& 125.10 & 6.30 & 18.95 & 5.68			
				& 102.41 & 6.77 & 20.07 & 6.04 
				& 170.58 & 5.36 & 15.71 & 4.97\\
				
				\rowcolor{rowgray} \cellcolor{white}
				& \textbf{NeuroKalman (Ours)}
				& \textbf{100.32} & \textbf{8.34} & \textbf{34.15} & \textbf{7.12}
				& \textbf{69.50} & \textbf{9.15} & \textbf{38.50} & \textbf{7.50}
				& \textbf{140.00} & \textbf{7.20} & \textbf{28.00} & \textbf{6.50}\\
				\bottomrule
			\end{tabular}
		\end{threeparttable}%
	} 
\end{table*}
%%%%%%%%%%%%%%%%%%%%%%%%%%%%%%%%%%%%%%%%%%%%%%%%%%%%%%%%%%%%%%%%%%%%%%%%%%%%%%%%%

\begin{table*}[t]
\centering
\caption{Low-data results on L1 Test-Seen. NeuroKalman reports mean$\pm$std over three 10\% trajectory-level random subsets. \textbf{Bold} indicates the best results.}
\label{tab:low_data}
\setlength{\tabcolsep}{5pt}
\renewcommand{\arraystretch}{.9}
\resizebox{\linewidth}{!}{
\begin{tabular}{l|l|cccc}
\toprule
\textbf{Methods} & \textbf{Protocol} & NE $\downarrow$ & SR $\uparrow$ & OSR $\uparrow$ & SPL $\uparrow$ \\
\midrule
TravelUAV~\cite{wang2024towards} & 100\% data & 106.28 & 16.10 & 44.26 & 14.30 \\
TravelUAV~\cite{wang2024towards} & 10\% w/o full pretrain & 122.63 & 13.19 & 37.16 & 11.71 \\
\rowcolor{rowgray}
\textbf{NeuroKalman (Ours)} & \textbf{10\% w/o full pretrain} & \textbf{85.89$\pm$2.71} & \textbf{22.70$\pm$0.87} & \textbf{45.04$\pm$1.09} & \textbf{19.04$\pm$0.70} \\
\bottomrule
\end{tabular}}
\end{table*}

\begin{table*}[t]
\caption{Effectiveness of fusion mechanism. All methods are implemented with the L1 assistant on the UAV-Need-Help test seen set. \textbf{Bold} indicates the best results among all methods.}
\label{tab:fusion}
\centering
\resizebox{\textwidth}{!}{
\setlength{\tabcolsep}{7pt}
\renewcommand{\arraystretch}{.9}
\begin{tabular}{l | cccc | cccc | cccc}
\toprule
\multirow{2}{*}{\textbf{Methods}} & \multicolumn{4}{c|}{\textbf{Full}} & \multicolumn{4}{c|}{\textbf{Easy}} & \multicolumn{4}{c}{\textbf{Hard}} \\
\cmidrule(lr){2-5} \cmidrule(lr){6-9} \cmidrule(lr){10-13}
& NE$\downarrow$ & SR$\uparrow$ & OSR$\uparrow$ & SPL$\uparrow$ & NE$\downarrow$ & SR$\uparrow$ & OSR$\uparrow$ & SPL$\uparrow$ & NE$\downarrow$ & SR$\uparrow$ & OSR$\uparrow$ & SPL$\uparrow$ \\
\midrule

$\mathbf{K}_t = 0.1$  & 217.09 & 0.00 & 0.00 & 0.00 & 132.81 & 0.00 & 0.00 & 0.00 & 302.56 & 0.00 & 0.00 & 0.00 \\
$\mathbf{K}_t = 0.5$  & 83.14 & 24.12 & 53.74 & 19.40 & 46.50 & 27.60 & 59.51 & 23.00 & 138.00 & 19.84 & 46.61 & 17.30 \\
$\mathbf{K}_t = 0.9$  & 100.96 & 18.05 & 44.15 & 15.35 & 58.18 & 19.92 & 52.11 & 18.43 & 154.50 & 15.80 & 34.33 & 14.50 \\
\rowcolor{rowgray} \textbf{Learnable} & \textbf{71.56} & \textbf{25.86} & \textbf{58.73} & \textbf{22.43} & \textbf{42.70} & \textbf{30.52} & \textbf{62.70} & \textbf{25.86} & \textbf{105.07} & \textbf{20.11} & \textbf{53.90} & \textbf{18.21} \\

\bottomrule
\end{tabular}}
\end{table*}

\subsection{Main Results}
\label{sec:main_results}
We conduct a comprehensive evaluation of NeuroKalman on the TravelUAV benchmark. Accordingly, we compare our method against several state-of-the-art baselines across two aspects: (1) Data Efficiency on the Test-Seen split, underscoring the model's robustness in long-horizon navigation; and (2) Generalization on the Test-Unseen splits (across both maps and objects), highlighting the model's capability to handle distribution shifts in unknown scenarios.

\noindent \textbf{Data Efficiency under Limited Fine-Tuning.}
Table~\ref{tab:seen} summarizes the Test-Seen results under the limited fine-tuning protocol. Specifically, NeuroKalman and TravelUAV-FT are initialized from full-data pretrained weights and then fine-tuned on the same fixed 10\% subset of the training data. Therefore, this comparison evaluates how effectively each model adapts with limited fine-tuning data. As shown, NeuroKalman achieves a significant performance lead across all metrics. For example, in the L1 Full split, our method attains a SR of 25.86\%, surpassing TravelUAV-FT (17.56\%) by a clear margin. This advantage is even more pronounced in the Hard split ($>250$m), where we improve the OSR from 36.85\% (TravelUAV-FT) to 53.90\% (Ours). This indicates that while the parametric TravelUAV-FT exhibits signs of overfitting and struggles to maintain trajectory consistency when starved of data, our method effectively mitigates this by integrating Kalman correction. \textbf{By anchoring the internal belief state to retrieved historical evidence, NeuroKalman prevents the progressive error accumulation.} Furthermore, our model consistently delivers superior performance when compared to TravelUAV. Notably, despite using only 10\% of the training data, NeuroKalman outperforms the TravelUAV on the L1 Hard split, reducing the NE from 152.04 to 105.07. This further demonstrates the robustness of NeuroKalman, validating that decoupling prediction and correction enables precise navigation.

\noindent \textbf{Data Efficiency without Full-Data Pretraining.}
The above limited fine-tuning protocol initializes both NeuroKalman and TravelUAV from weights pre-trained on the full training set. To verify that the improvement is not merely inherited from full-data pretraining, we further train the model without full-data pretraining using only 10\% of TravelUAV training trajectories. The subset is uniformly sampled at the trajectory level with fixed random seeds, and the sampled splits preserve the Easy/Hard distribution of the full set (51.8\%/48.2\% on average vs. 52.4\%/47.6\% in the full training set).

As shown in Table~\ref{tab:low_data}, NeuroKalman substantially outperforms TravelUAV under the 10\% low-data setting without full-data pretraining, reducing NE by 30.0\% and improving SR by 72.1\%. It also remains competitive with, and in most metrics surpasses, the fully trained TravelUAV baseline, despite using only 10\% training trajectories. Meanwhile, TravelUAV only predicts waypoints step-by-step from current observations without explicit long-horizon error correction, leaving its performance heavily bottlenecked by the MLLM’s inherent capacity and resulting in similarly low performance under both 10\% and 100\% data. This suggests that \textbf{NeuroKalman breaks this bottleneck via explicit error correction and temporal modeling, demonstrating its low-data efficiency.}

\begin{table*}[t]
\caption{Impact of memory history length. All methods are implemented with the L1 assistant on the UAV-Need-Help test seen set. \textbf{Bold} indicates the best results among all methods.}
\label{tab:memory}
\centering
\resizebox{\linewidth}{!}{ 
\setlength{\tabcolsep}{7pt}
\renewcommand{\arraystretch}{1}

\begin{tabular}{l | cccc | cccc | cccc}
\toprule

\multirow{2}{*}{\textbf{Methods}} & \multicolumn{4}{c|}{\textbf{Full}} & \multicolumn{4}{c|}{\textbf{Easy}} & \multicolumn{4}{c}{\textbf{Hard}} \\

 & NE$\downarrow$ & SR$\uparrow$ & OSR$\uparrow$ & SPL$\uparrow$
 & NE$\downarrow$ & SR$\uparrow$ & OSR$\uparrow$ & SPL$\uparrow$
 & NE$\downarrow$ & SR$\uparrow$ & OSR$\uparrow$ & SPL$\uparrow$ \\
\midrule

$M=5$ & 84.39 & 21.23 & 53.46 & 18.18
      & 51.56 & 25.03 & 56.96 & 21.21
      & 125.00 & 16.54 & 49.13 & 15.09 \\

\rowcolor{rowgray}
$M=10$ & \textbf{71.56} & \textbf{25.86} & \textbf{58.73} & \textbf{22.43}
       & \textbf{42.70} & \textbf{30.52} & \textbf{62.70} & \textbf{25.86}
       & \textbf{105.07} & \textbf{20.11} & \textbf{53.90} & \textbf{18.21} \\

$M=15$ & 77.17 & 23.77 & 56.42 & 20.16
       & 47.75 & 28.10 & 60.15 & 23.33
       & 116.21 & 18.43 & 51.81 & 17.39 \\

\bottomrule
\end{tabular}
}
\end{table*}

\noindent \textbf{Generalization on the Test-Unseen Split.}
We further evaluate the model's generalization by testing on the Test-Unseen-Object and Test-Unseen-Map splits. As summarized in Table~\ref{tab:unseen}, NeuroKalman consistently outperforms baseline methods~\cite{anderson2018vision, wang2024towards} in these novel settings on the Full split. For example, on the Unseen Objects (UO) split, our method achieves a SR of 32.48\% and an OSR of 60.82\%, significantly surpassing the strong NavFoM baseline (SR: 29.83\%, OSR: 47.99\%). The performance gap is even more evident in the challenging Unseen Maps (UM) split, where NeuroKalman nearly doubles the SR of TravelUAV (8.34\% vs. 4.18\%). While NavFoM achieves a slight advantage in SR and SPL on the Hard split of unseen objects due to its large-scale data pre-training, NeuroKalman maintains a considerably lower NE (84.50 vs. 133.01) and a higher OSR (58.00\% vs. 46.18\%). These results highlight the limitation of purely parametric baselines, which suffer from performance degradation due to error accumulation caused by state drift when facing unknown topologies or objects. In contrast, the generalization of NeuroKalman validates the efficacy of Bayesian fusion mechanism. \textbf{By adaptively regulating the reliance between the GRU’s kinematic prior and the measurement likelihood, our framework successfully utilizes retrieval-based anchors to correct accumulated errors.}

%%%%%%%%%%%%%%%%%%%%%%%%%%%%%%%%%%%%%%%%%%%%%%%%%%%%%%%%%%%%%%%%%%%%%%%%%%%%%%%%%

\subsection{Ablation Study}

\noindent \textbf{Effectiveness of Fusion Mechanism.}
Next, we investigate whether a learnable Kalman Gain is superior to fixed scalar values in balancing the motion prior and measurement evidence. As presented in Table~\ref{tab:fusion}, fixed strategies fail to achieve optimal navigation results. A strong bias towards the Prior ($\mathbf{K}_t = 0.1$) leads to catastrophic failure. Conversely, relying heavily on the Measurement ($\mathbf{K}_t = 0.9$) also yields suboptimal performance, with the SR dropping to 18.05\% on the Full split compared to the balanced setting. Even the best fixed strategy ($\mathbf{K}_t = 0.5$) is inferior to our approach. In contrast, our learnable gating mechanism consistently achieves superior performance across all metrics, reaching the highest SR of 25.86\% and the lowest NE of 71.56. We think that the failure at $\mathbf{K}_t = 0.1$ confirms that without sufficient external correction, the model suffers from unbounded state drift. Meanwhile, the degradation at $\mathbf{K}_t = 0.9$ suggests that ignoring temporal smoothness makes the model vulnerable to noisy retrievals. Our learnable gating network effectively acts as an uncertainty switch, which dynamically \textbf{integrates internal dynamics and external measurements to maintain robust estimation across diverse noise regimes.}

\noindent \textbf{Impact of Memory History Length.}
We also investigate the impact of memory history length $M$ to determine the optimal temporal context for our retrieval mechanism. As shown in Table~\ref{tab:memory}, deviations from our design choice lead to distinct performance drops. A short history ($M=5$) yields suboptimal results, with a higher NE of 84.39 on the Full split. Interestingly, extending the history excessively ($M=15$) does not help but rather degrades performance, increasing the NE to 77.17. Our setting ($M=10$) achieves the best performance with the lowest NE of 71.56. We think that a limited temporal window lacks sufficient historical anchors to recover from accumulated drift, whereas an excessive length introduces outdated visual features that act as noise, distracting the attention mechanism from relevant evidence. This indicates that \textbf{appropriate memory length ensures sufficient context for effective re-localization without introducing spurious correlations.}

%%%%%%%%%%%%%%%%%%%%%%%%%%%%%%%%%%%%%%%%%%%%%%%%%%%%%%%%%%%%%%%%%%%%%%%%%%%%%%%%%

\noindent \textbf{Additional Robustness Checks.}
We further evaluate NeuroKalman's robustness to output-space smoothing and memory noise by introducing a post-hoc Kalman filtering baseline and varying the memory insertion threshold. On the one hand, we compare against a post-hoc Kalman baseline that applies a constant-velocity Kalman filter to TravelUAV's predicted waypoints. As shown in Table \ref{tab:post_kf}, this output-space smoothing improves TravelUAV only mildly on L1 Test-Seen (NE/SR: 106.28/16.10 $\rightarrow$ 96.67/18.17), still far behind NeuroKalman (71.56/25.86), showing that \textbf{latent-space correction with semantic memory cannot be replaced by geometric smoothing alone.} On the other hand, we vary the memory insertion threshold $\sigma_t \in \{0.3,0.5,0.7,0.9\}$. As shown in Table \ref{tab:threshold}, NeuroKalman remains stable for $\sigma_t \ge 0.5$, with $\sigma_t = 0.5$ yielding the best performance. A low threshold ($\sigma_t = 0.3$) hurts performance by admitting noisy anchors. Our method exhibits a robust threshold range: memory is inserted only through post-correction selective storage, and the posterior is filtered via confidence-aware Kalman fusion, ensuring that noisy retrievals are not unconditionally trusted.

\begin{figure}[t!]
	\centering
	\includegraphics[width=1\linewidth]{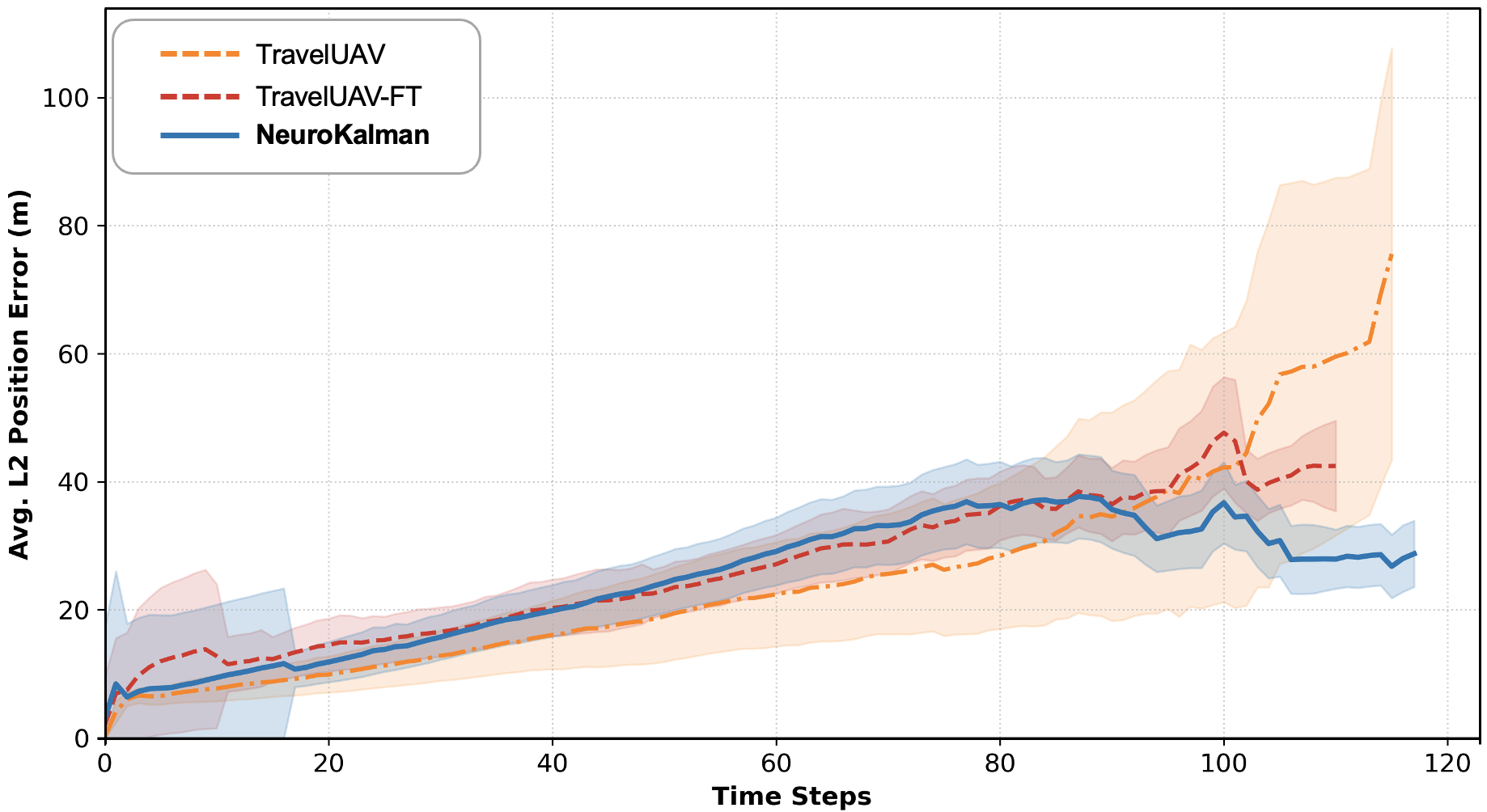}
	%\vspace{-4mm}
	\caption{\textbf{Visualization of $L_2$ position error over time.} The baselines (orange and red dashed lines) show a continuous error increase on long trajectories. Conversely, NeuroKalman (blue solid line) keeps the error stable and prevents it from growing rapidly via effective Kalman correction.}
	\label{fig:error}
	%\vspace{-4mm}
\end{figure}

\begin{table}[t]
\centering
\caption{Comparison with post-hoc Kalman filtering on L1 Test-Seen set. \textbf{Bold} indicates the best results.}
\label{tab:post_kf}
\begingroup
\setlength{\tabcolsep}{5pt}
\renewcommand{\arraystretch}{1.2}
\resizebox{\linewidth}{!}{
\begin{tabular}{l|cccc}
\toprule
\textbf{Methods} & NE $\downarrow$ & SR $\uparrow$ & OSR $\uparrow$ & SPL $\uparrow$ \\
\midrule
TravelUAV & 106.28 & 16.10 & 44.26 & 14.30 \\
TravelUAV + Post-KF & 96.67 & 18.17 & 33.65 & 15.64 \\
\rowcolor{rowgray}
\textbf{NeuroKalman (Ours)} & \textbf{71.56} & \textbf{25.86} & \textbf{58.73} & \textbf{22.43} \\
\bottomrule
\end{tabular}}
\endgroup
\end{table}

\begin{table}[t]
\centering
\caption{Sensitivity to memory insertion threshold on L1 Test-Seen set. \textbf{Bold} indicates the best results.}
\label{tab:threshold}
\begingroup
\setlength{\tabcolsep}{10pt}
\renewcommand{\arraystretch}{1}
\resizebox{\linewidth}{!}{
\begin{tabular}{c|cccc}
\toprule
\textbf{Threshold} & NE $\downarrow$ & SR $\uparrow$ & OSR $\uparrow$ & SPL $\uparrow$ \\
\midrule
$\sigma_t=0.3$ & 82.45 & 20.50 & 48.15 & 17.22 \\
\rowcolor{rowgray}
$\sigma_t=0.5$ & \textbf{71.56} & \textbf{25.86} & \textbf{58.73} & \textbf{22.43} \\
$\sigma_t=0.7$ & 73.88 & 24.55 & 55.62 & 22.10 \\
$\sigma_t=0.9$ & 75.45 & 24.18 & 53.30 & 21.75 \\
\bottomrule
\end{tabular}}
\endgroup
\end{table}

%%%%%%%%%%%%%%%%%%%%%%%%%%%%%%%%%%%%%%%%%%%%%%%%%%%%%%%%%%%%%%%%%%%%%%%%%%%%%%%%%
\subsection{Drift Analysis}
\label{sec:drift_analysis}
Finally, we analyze the state drift problem by visualizing the average $L_2$ position error over time in Figure~\ref{fig:error}. Intuitively, the baseline curves (TravelUAV and TravelUAV-FT) terminate earlier than NeuroKalman because baseline models are more prone to premature collisions or severe drift on long trajectories. In detail, their localization errors accumulate over time and become especially evident after 100 steps. In contrast, the error of NeuroKalman stops growing and stays stable (around 30--40 meters) after a small initial rise. This demonstrates that our Kalman correction mechanism successfully \textbf{leverages retrieved memory anchors to periodically rectify the internal belief}, ultimately constraining error accumulation in long-horizon navigation.

%%%%%%%%%%%%%%%%%%%%%%%%%%%%%%%%%%%%%%%%%%%%%%%%%%%%%%%%%%%%%%%%%%%%%%%%%%%%%%%%%

%\vspace{-2mm}
\section{Conclusion}
%\vspace{-2mm}
\label{sec:conclus}
In this work, we reframe the continuous navigation task as a Recursive Bayesian State Estimation problem to address the fundamental challenge of state drift inherent in open-loop parametric inference. We propose NeuroKalman, a framework that decouples navigation into prior prediction via motion dynamics and likelihood correction via memory anchors, while mathematically establishing the equivalence between attention-based retrieval and Kernel Density Estimation (KDE). Comprehensive experiments on TravelUAV benchmark demonstrate that our method achieves remarkable data efficiency and generalization, clearly outperforming strong baselines and regulating drift accumulation.

\textbf{Limitation.} The current implementation of the prediction prior utilizes a GRU-based RNN as baseline, which may be subject to information decay over exceptionally long horizons. However, the primary contribution of NeuroKalman is recursive Bayesian correction mechanism and we can easily apply another architecture to enhance the overall robustness.

\section*{Acknowledgement}
This work was supported in part by the National Natural Science Foundation of China (NSFC) Young Scientists Fund (Type B) under Grant No. 62522220; in part by the NSFC General Program under Grant No. 62172439, and in part by the NSFC General Program under Grant No. 62502405.

\section*{Impact Statement}
This work advances robust UAV navigation in GPS-denied environments, with potential for search-and-rescue and disaster relief. Although effective, rigorous verification processes and safety protocols are strictly necessary prior to integrating it into real-world embodied systems.

\nocite{langley00}
% \newpage
{
\balance
\bibliography{icml2026bib}
\bibliographystyle{icml2026}
}

%You can have as much text here as you want. The main body must be at most $8$ pages long.
%For the final version, one more page can be added.
%If you want, you can use an appendix like this one.  

%The $\mathtt{\backslash onecolumn}$ command above can be kept in place if you prefer a one-column appendix, or can be removed if you prefer a two-column appendix.  Apart from this possible change, the style (font size, spacing, margins, page numbering, etc.) should be kept the same as the main body.
%%%%%%%%%%%%%%%%%%%%%%%%%%%%%%%%%%%%%%%%%%%%%%%%%%%%%%%%%%%%%%%%%%%%%%%%%%%%%%%
%%%%%%%%%%%%%%%%%%%%%%%%%%%%%%%%%%%%%%%%%%%%%%%%%%%%%%%%%%%%%%%%%%%%%%%%%%%%%%%

%%%%%%%%%%%%%%%%%%%%%%%%%%%%%%%%%%%%%%%%%%%%%%%%%%%%%%%%%%%%%%%%%%%%%%%%%%%%%%%
%%%%%%%%%%%%%%%%%%%%%%%%%%%%%%%%%%%%%%%%%%%%%%%%%%%%%%%%%%%%%%%%%%%%%%%%%%%%%%%
% APPENDIX
%%%%%%%%%%%%%%%%%%%%%%%%%%%%%%%%%%%%%%%%%%%%%%%%%%%%%%%%%%%%%%%%%%%%%%%%%%%%%%%
%%%%%%%%%%%%%%%%%%%%%%%%%%%%%%%%%%%%%%%%%%%%%%%%%%%%%%%%%%%%%%%%%%%%%%%%%%%%%%%
\newpage
\appendix
\onecolumn

% \section{Appendix}
% %\vspace{-2mm}
% \label{sec:appendix}
% \input{sec/appendix} 

\end{document}